# MeshConv3D: Efficient convolution and pooling operators for triangular 3D meshes

Germain Bregeon
*Keyrus*
*Telecom SudParis*
*Institut Polytechnique de Paris*
germain.bregeon@telecom-sudparis.eu

Marius Preda
*Telecom SudParis*
*Institut Polytechnique de Paris*
marius.preda@telecom-sudparis.eu

Radu Ispas
*Keyrus*
radu.ispas@keyrus.com

Titus Zaharia
*Telecom SudParis*
*Institut Polytechnique de Paris*
titus.zaharia@telecom-sudparis.eu

*Abstract*— Convolutional neural networks (CNNs) have been pivotal in various 2D image analysis tasks, including computer vision, image indexing and retrieval or semantic classification. Extending CNNs to 3D data such as point clouds and 3D meshes raises significant challenges since the very basic convolution and pooling operators need to be completely re-visited and re-defined in an appropriate manner to tackle irregular connectivity issues. In this paper, we introduce MeshConv3D, a 3D mesh-dedicated methodology integrating specialized convolution and face collapse-based pooling operators. MeshConv3D operates directly on meshes of arbitrary topology, without any need of prior re-meshing/conversion techniques. In order to validate our approach, we have considered a semantic classification task. The experimental results obtained on three distinct benchmark datasets show that the proposed approach makes it possible to achieve equivalent or superior classification results, while minimizing the related memory footprint and computational load.

## I. INTRODUCTION

Convolutional layers hold a crucial role in neural networks when applied to 2D image analysis. They efficiently extract local, translation invariant features that are subsequently used for robust and salient pattern detection. When combined with pooling layers, they make it possible to jointly reduce the dimensionality of the representation and enhance its generalization capabilities.

Extending convolutional operators to 3D data, such as volumetric images, 3D point clouds and 3D meshes is an important challenge, due to the potential of 3D convolution features to capture rich spatial information.

The specification of such fully 3D representations is essential for various application domains that involve accurate and effective 3D object detection and recognition capabilities. Among them, let us mention autonomous driving and robotics, where the object detection and recognition accuracy is a high priority, virtual and augmented reality, where the precise understanding of the 3D environment is essential for creating immersive and interactive experiences or medical imaging, where the analysis and identification of specific 3D patterns is of crucial importance.

The two mostly used 3D data representations are today point clouds and meshes.

The primary challenge in the case of point clouds is related to the unstructured character of such representations, which completely lacks of any topological information. The difficulty then is to consistently define the convolutional regions and corresponding weights over all the points in the cloud.

In the case of meshes, the definition of a convolution kernel is facilitated by the availability of the inherent neighboring information provided by edges. However, the main difficulty in this case is related to the highly irregular nature of the mesh topology, which makes it difficult to define consistent convolution kernels.

In this paper, we introduce the so-called MeshConv3D approach, which integrates two distinct contributions:
- a specialized convolutional operator specifically designed for triangular meshes of arbitrary topology,
- a highly efficient pooling mechanism optimized in both memory footprint and computational complexity.

MeshConv3D operates directly on raw meshes without requiring any remeshing process and incorporates as face features the two descriptors from ExMeshCNN [1]. The convolution regions are established using a translational and rotational invariant ordering method that can be flexibly adjusted to accommodate variable sizes of the receptive field. Pooling is then efficiently performed by a dedicated face collapse operator, which makes it possible to eliminate in a single pass approximately half of the mesh faces.

The rest of the paper is organized as follows. Section II presents an overview of the state of the art and analyzes the limitations of existing methods.

Section III introduces the proposed MeshConv3D approach and details feature specification, convolutional and pooling operators. Section IV summarizes the experimental results obtained for mesh classification tasks on the SHREC11 [2], cubes [3] and Manifold40 [4] datasets. Finally, Section V concludes the paper and opens some perspectives of future work.

## II. STATE OF THE ART

In recent years, the scientific community has considered the challenging issue of developing deep learning models specifically tailored for 3D model analysis purposes.

Early efforts in this field involve 2D/3D techniques [5], [6], where the 3D model is represented as a set of 2D projections. Such an approach offers the advantage of enabling 3D model analysis through well-established networks, operating over the 2D domain. However, they are heavily reliant on the critical projection step, and vulnerable to variations in pose.

On the other hand, point cloud analysis offers a compelling avenue for studying the geometric intricacies of 3D models directly in their native domain.

PointNet [7] was among the pioneering neural networks specifically designed for feature extraction from point clouds and employs a multilayer perceptron architecture.

Other convolutional operators, such as SparseCNN [8] or RS-CNN [9] cater to different forms of 3D data, such as voxelized point clouds and leverage various types of 3D convolutional layers. The sparse 3D convolution layers [8] represent a notable breakthrough in this regard. By discretizing the input point cloud into a grid of voxels and focusing the convolution operations solely on the occupied voxels, such approaches effectively overcome the challenge of defining neighborhoods in a sparse 3D space.

Similarly, the RS-CNN approach [9] delineates local neighborhoods around specific points, leveraging fundamental relationships to comprehensively map and understand the intricate structure of the data. The convolutional layers involved demonstrate enhanced efficiency during the learning process and offer a powerful tool to capturing both local and global patterns within the data, as well as spatial relationships.

In the case of 3D meshes, the connectivity information, which defines the mesh topology, is almost always highly irregular, with a number of neighboring vertices or faces that can significantly vary from a mesh region to another. This represents the main challenge that needs to be overcome when designing dedicated convolutional operators.

An early solution, introduced in [10], converts the mesh connectivity to a regular one with the help of a re-meshing technique, in order to ensure that each vertex has precisely six neighbors. This enables the subsequent application of 2D hexagonal convolutional operators [11].

The GCNN method [12] operates by selecting patches of vertices based on their relative positions, ensuring the capture of significant shape features.

In contrast, DiffusionNet [13] adopts a different strategy, utilizing a pointwise function, represented as a Multilayer Perceptron (MLP) in the mesh spectral domain. The technique makes it possible to transform features across vertices and diffuse the information among neighboring points, while capturing local data changes.

The MeshWalker model [14] delves into mesh connectivity and topology by generating random walks as lists of vertices, which are then fed into a recurrent neural network to extract mesh features.

While such vertex-based methods can effectively handle straightforward mesh tasks, they solely rely on vertex information, potentially overlooking certain geometric intricacies.

Another widely adopted convolutional operator, so-called MeshCNN [3], defines the convolution operation based on edge features. The convolution kernel consists of a central edge and its four neighbors forming the adjacent triangles (under the assumption of manifold meshes), facilitating mesh dimension reduction through a contraction operation on the selected edge. The features used remain invariant to rotation and translation and can be fully reconstructed.

PD-MeshNet [15] employs two types of graph features for edges and faces, aggregating them through a graph neural network along with an attention mechanism. Here, the pooling operation is executed by edge contraction on a selected edge.

However, such convolutional approaches rely on fixed kernel sizes, potentially limiting the receptive field of the convolution operation. Moreover, the proposed pooling operations can be inefficient, as each edge pooling requires re-computation to determine the next contracted edge, an operation which is not parallelizable. In contrast, our new convolution operator supports varying kernel sizes, and the proposed pooling mechanism simultaneously eliminates a significant number of faces.

In [16], each mesh face serves as a fundamental element for convolution. Intrinsic face features, such as dihedral angles, face areas and internal angles compose the input to the convolutional layer. Convolution regions for each face are established by sorting mesh faces with a 1D convolutional kernel, and then iteratively incorporating new neighbors until the desired region size is achieved. The method enables the application of a standard convolution operator on the constructed regions, with pooling accomplished through face collapse.

The SubdivNet [4] approach re-triangulates each mesh so that to conform to a Loop subdivision structure, thus greatly simplifying kernel selection and pooling processes. However, the remeshing process significantly increases the number of mesh faces, leading to prohibitive memory requirements.

MeshNet++ [17] and ExMeshCNN [1] introduce two descriptor layers based on intrinsic face features: a geodesic descriptor capturing the global face position within the mesh, and a geometric descriptor capturing the face's local shape and its relationship with the neighbors. However, such solutions are penalized by the lack of an appropriate feature aggregation mechanism. These methods achieve state-of-the-art results on benchmark mesh classification tasks.

The convolutional operator proposed in this paper shares similarities with the approaches proposed by SubdivNet [4] and ExMeshCNN [1]. However, it offers the advantage of supporting varying region sizes, a feature not supported by ExMeshCNN.

In addition, unlike ExMeshCNN [1], our solution incorporates a feature aggregation step, which consists of a pooling algorithm. Let us underline that the pooling mechanism facilitates dimensionality reduction, enhances computational efficiency, and prevents overfitting by integrating data from neighboring regions. Moreover, it helps the feature selection process by prioritizing salient features within regions, thus promoting generalization, and facilitating robust learning.

SubdivNet [4] introduces a pooling algorithm based on a Loop subdivision scheme, which necessitates a complete remeshing. However, neither of the remeshing algorithms retained achieves satisfactory results: the MAPS [18] algorithm occasionally introduces visible distortions, while the refined maps algorithm [19] struggles to achieve low resolution sizes in the case of complex input meshes. Moreover, the remeshing process significantly increases the complexity of the network in terms of memory usage, multiplying the number of faces in each mesh by up to 40 times.

In contrast, the MeshConv3D solution proposed in this paper operates on native meshes without requiring any preprocessing. In addition, the proposed pooling algorithm demonstrates superior efficiency in terms of memory usage and time, thanks to its parallelized structure, compared to both face-based methods and MeshNet++ [17]. Furthermore, our approach achieves comparable results on mesh classification tasks (*cf*. Section IV).

Let us now detail the proposed methodology.

### III. MESHCONV3D APPROACH

We define a triangular mesh in a simplified form $\mathcal{M} = (\mathcal{V}, \mathcal{F}, \mathcal{E})$ as a set of vertices $\mathcal{V} = \{v_i | v_i \in \mathbb{R}^3\}$, a set of edges $\mathcal{E} = \{e_i | e_i \in \mathcal{V}^2\}$ and a set of triangular faces $\mathcal{F} = \{f_i | f_i \in \mathcal{V}^3\}$. Let $V$, $E$ and $F$ respectively denote the number of vertices, edges and faces.

We deliberately disregard in this representation the ordering/orientation information but assume that we have to deal with consistently oriented manifold meshes.

The adjacency matrix $A_\mathcal{M}$ is a size ($F \times 3$) matrix whose $i^{th}$ row stores the integers indices of the three neighboring faces to face $f_i$. By definition, two faces are considered as neighbors if they share a common edge. If a face has less than three neighbors, which can appear in the case of meshes with borders, we simply add a zero value. The neighboring faces in the adjacency matrix are sorted according to the length of their edge shared with the central face.

Fig. 1 illustrates an example of adjacency matrix for a simple mesh (solely rows corresponding to faces A and C are here explicited).

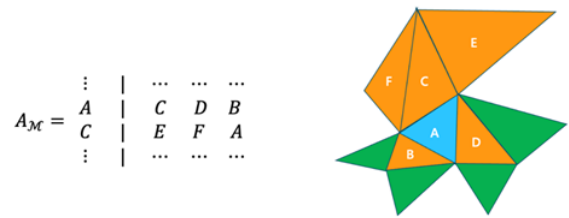

Figure 1: (right) Convolution region of face A, (left) adjacency matrix of faces A and C.

### A. Local face descriptors

We consider as local face descriptors the empirical geodesic and geometric features introduced in [1] and recalled here below.

For a given face $f = (v_f^1, v_f^2, v_f^3) \in \mathcal{F}$, of gravity center $c_f$, the geodesic descriptor $K_f$ takes as input a collection of intrinsic features and is defined as:

$$K_f = \alpha_0 \sum_{i=1}^{3}(v_f^i - c_f) + \alpha_1 \sum_{i=1}^{3} n_f^i +$$
$$\alpha_2 \sum_{i=1}^{3} | e_f^{i,1} - e_f^{i,2} | \quad , \tag{1}$$

where $n_f^i$ denotes the normal vector to vertex $v_f^i$, defined as the average normal of its incident faces, and $e_f^{i,1/2}$ denotes the two vectors formed by the incident edges to vertex $v_f^i$ belonging to the corresponding adjacent triangles. The last term in equation (1) expresses the relationship between two geodesic paths passing through the same vertex of the central face. The parameters $\alpha_i$ are learnable weights that can be assimilated to a simplified 1D convolutional kernel.

The geometric descriptor $G_f$ learns high-level geometric features from low level intrinsic ones and is defined as described in the following equation:

$$G_f = \beta_0 c_f + \beta_1 n_f + \beta_2 \sum_{i=1}^{3} |c_f - c_f^i|$$
$$+\beta_3 \sum_{i=1}^{3} |n_f \times n_f^i|, \quad (2)$$

where $n_f$ is the face $f$ normal vector, $c_f$ and $c_f^j$ respectively denote the centroids of the central face and of its neighboring faces. Let us note that the outer product $n_f \times n_f^i$ represents the angle formed by two adjacent faces. Similarly to the geodesic descriptor, the learnable weights $\beta_i$ are equivalent to a 1D convolution filter.

The first two terms in equation (2) learn the position and orientation of the target face in the mesh, while the remaining two capture the relationship between the target face and its neighbors in terms of geometric distances and angles.

The final face descriptor $d_f$ is simply obtained as the concatenation of the geodesic and geometric features:
$$d_f = (K_f, G_f) \quad (3)$$

The $d_f$ descriptor is subsequently used as input to the proposed convolution operator described in the following section.

### B. Mesh convolution layer

In order to define the proposed convolution operation, we start by specifying a set of local patches for each face of the mesh, denoted by $R_{F \times K}$, which serve as region of support for the convolution kernel. The integer parameter $K$ corresponds to the desired convolutional kernel size (expressed as the number of surrounding triangles). For a given face, the local patch is initialized with itself together with its three adjacent faces. The resulting set of patches is $R_{F \times 4}$.

To expand the receptive fields, additional adjacent faces are added to $R$, by sequentially traversing the adjacent faces and incrementally including their corresponding neighbors. This process continues iteratively until the desired number of faces $K$ is reached.

For a given face $f$, the proposed convolution operation is finally defined as:

$$Conv(d_f) = w_0 d_f + w_1 \sum_{n=1}^{K} d_f^n +$$
$$w_2 \sum_{n=1}^{K} |d_f - d_f^n| \quad (4)$$

Here, $d_f$ represents the central face's descriptor, while $d_f^n$, $n = 1, \dots, K$ denote the descriptors of the other $K$ faces within this region. The parameters $w_0$, $w_1$ and $w_2$ are learnable weights.

We have opted for this operation to address the face ordering issue inherent in mesh structures. In a classical mesh convolution operation, determining the correct face order can be ambiguous. However, by employing a sum that is ordering invariant, we circumvent the need to select a specific face order within the convolution region. This operation enables us to establish a user-defined convolution region, thereby allowing the network to have a broader receptive field.

Let us now detail the second essential step of our approach, which concerns the pooling layer.

### C. Mesh pooling layer

The proposed pooling layer is based on a face-collapse operation. We first assign a weight to each face $f$ of the mesh, denoted as $w_f$, that captures the local level of significance of the face in its neighborhood. This weight is defined as the L$_2$ distance between the features of the considered face and those of its three neighboring triangles ($d_f^n$):

$$w_f = \sum_{n=1}^{3} \|d_f - d_f^n\|^2 \quad (5)$$

Considering a target pooling size $T$ (expressed as number of remaining mesh faces after pooling), the pooled mesh is constructed as follows:

1. For each face, a candidate pooling region is defined by itself and its three neighboring faces, along with the faces whose connectivity would be altered by their removal (Fig. 2).
2. Select the face $f$ with the lowest weight $w_f$ over the entire mesh.
3. Mark all faces whose pooling is incompatible with the selected face, i.e. faces whose pooling regions overlap with the one of the selected face.
4. While the target pooling size $T$ is not reached, iterate steps 2 and 3 on the remaining set of unmarked faces not yet selected for pooling.
5. Simultaneously perform the pooling for all the pooling regions determined (Fig. 3) and update accordingly the new mesh adjacency matrix. The features associated to the resulting pooled faces are defined as the average value of its neighbors in the corresponding pooling region.
6. At the end of step 5, we obtain a pooled mesh with a size that still may be superior to the target size $T$. If this is the case, then repeat steps 1 to 5 until the target size $T$ is achieved. Otherwise, Stop the pooling algorithm.

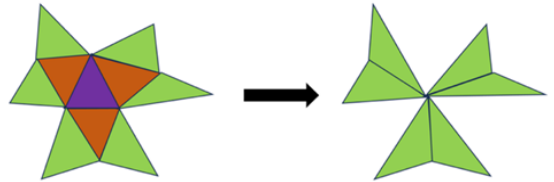

Figure 2: Example of pooling of a selected face; in purple, the selected face, in purple and orange, the faces that will disappear after the pooling operation, in green the faces whose connectivity is modified by the pooling operation.

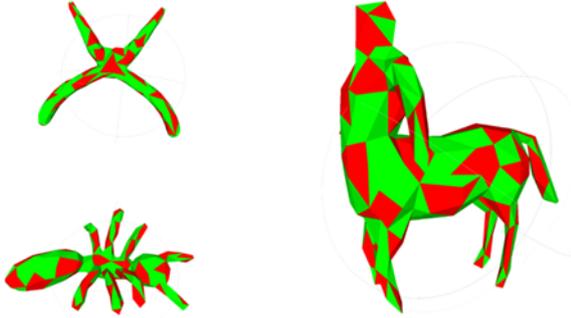

Figure 3: Pooling iteration: in red, the faces that are removed by the operation and in green the faces that remain after pooling.

The proposed algorithm allows for parallel computation of all mesh pooling regions, resulting in the simultaneous removal of a significantly high (in most cases, about a half) number of faces at each iteration. In addition, the operations have been tailored to handle batches of meshes, enhancing their suitability for deep-learning environments.

## IV. EXPERIMENTAL RESULTS

To validate the proposed methodology, we have considered a semantic classification task, performed on various publicly available datasets.

### A. Network architecture

We have adopted a popular 2D convolutional architecture to leverage the regularity afforded by our newly developed operations. We have notably selected a VGG-like network for our classification task. This network comprises 3 blocks of convolution layers each followed by a mesh pooling layer, and a final global average pooling layer to obtain the score for each class. We set the pooling parameter $T$ to 400, 300, and 200 for each respective pooling operation. This architecture is very similar to the one used by other state of the art models such as MeshCNN [3], ExMeshCNN [1] or SubdivNet [4] and thus ensures a fair comparison.

### B. Datasets for mesh classification

Our experiments have been carried out on three distinct datasets widely used in the state of the art for benchmarking mesh classification methods: SHREC11, cubes and Manifold40.

SHREC11 [2] is a dataset comprising 30 classes, with 20 meshes per class. Each mesh in the dataset is composed of 500 faces. We adhere to the setting established in MeshCNN [3] to enable subsequent comparisons of our work with various state-of-the-art models by creating two different splits. The first method involves randomly selecting 16 models from each class in the training set and utilizing the remaining 4 models for testing (16-split). The second one randomly selects 10 models from each class for the training set, while the remaining 10 models are incorporated into the testing set (10-split).

The cube engraving dataset [3] consists of cubes adorned with patterns extracted from 22 classes of the MPEG 7 binary shape dataset [20]. It comprises 3722 models in the training set and 659 models in the testing set. All the models are composed of 500 faces.

The last dataset retained is Manifold40 [4], which comprises exclusively closed manifold meshes, each containing 500 faces, derived from the ModelNet40 [21] corpus. The meshes are categorized into 40 classes, with 9843 meshes for training and 2468 for testing.

### C. Data pre-processing

We apply the most widely used pre-processing method in the literature. As all meshes are located in different positions and have different scales, we relocate the center point of each mesh to the origin and normalize it within a unit sphere.

Subsequently, we compare our results with those of various state-of-the-art models.

### D. Classification accuracy results

Table I presents the results obtained on the two different setups of SHREC11 as well as on the cubes engraving dataset.

TABLE I. CLASSIFICATION ACCURACY ON THE SHREC11 [13] AND CUBES [3] DATASETS.

| Method | SHREC11 | | cubes |
| --- | --- | --- | --- |
| | Split 10 | Split 16 | |
| MeshCNN [3] | 91,0% | 98,6% | 92,2% |
| PD-MeshNet [15] | 99,1% | 99,7% | 94,4% |
| MeshWalker [14] | 97,1% | 98,6% | 98,6% |
| MeshNet++ [17] | 100% | 99,8% | 98,5 |
| SubdivNet [4] | 100% | 100% | 100% |
| Face-based [16] | 100% | 100% | 99,4% |
| ExMeshCNN [1] | 100% | 100% | 100% |
| **MeshConv3D** | **100%** | **100%** | **100%** |

Comparing to other state-of-the-art results, we observe that on SHREC11, our model achieves state-of-the-art performance by achieving perfect accuracy on the testing set in both settings.

On the cubes dataset, MeshConv3D also achieves state-of-the-art performance.

Table II summarizes the accuracy results of MeshConv3D alongside several state-of-the-art models. MeshConv3D falls within the range of state-of-the-art performances. The results are higher than those achieved by its closest competitor, SubdivNet, but without the increased complexity related to the re-meshing process required by SubdivNet, which leads to a significant higher number of faces. On this corpus, ExMeshCNN achieves better results than MeshConv3D. However, it lacks a general and efficient pooling layer for triangular meshes, making it less generic on meshes with a large number of faces, since pooling reduces the dimensionality of data, enhances computational efficiency, promotes model generalization, and facilitates the learning of robust features by summarizing information from neighboring features.

TABLE II. CLASSIFICATION ACCURACY ON THE MANIFOLD40 [4] DATASET.

| Method | Manifold40 |
|---|---|
| MeshNet [22] | 88,4% |
| MeshWalker [14] | 90,5% |
| SubdivNet [4] | 91,4% |
| ExMeshCNN [1] | 93,6% |
| **MeshConv3D** | **92,4%** |

*E. Performance review*

The criteria retained for evaluating the performances are the runtime ratio and peak memory usage. The runtime ratio, quantified as the ratio between the execution time of alternative methods and the one of our approach, alongside peak memory usage—representing the maximum GPU memory required during training in megabytes (MB) have been assessed. These metrics are derived from an average of 50 batches, each comprising 50 meshes of the SHREC11 dataset, encompassing both forward and backward pass phases. Experiments have been conducted using an NVIDIA GeForce RTX 2080 Ti GPU with 12 GB of memory, with no interfering processes. The findings of this investigation furnish insights into the comparative performance of diverse mesh convolution techniques, thereby offering critical guidance for the refinement and optimization of neural network architectures tailored to mesh data processing.

Table III summarizes the efficiency results of MeshConv3D compared to three state of the art methods. In comparison to our implementation, both MeshCNN and MeshNet++ exhibit slower performance and significantly higher GPU memory consumption.

Although SubdivNet demonstrates a slightly faster runtime than our method, its requirement for remeshing to achieve face organization regularity leads to a substantial increase in face count compared to the original meshes of the dataset (up to 30 to 40 times more faces in the remeshed dataset). Consequently, this imposes a higher demand on GPU memory, which is approximately 10 times greater than our technique.

TABLE III. RUNTIME RATIO AND PEAK MEMORY USAGE

| Method | Runtime ratio | Peak memory usage (MB) |
|---|---|---|
| MeshCNN [3] | 1,40 | 2596 |
| MeshNet++ [17] | 3,81 | 7440 |
| SubdivNet [4] | 0,87 | 6981 |
| **MeshConv3D** | **1** | **682** |

These findings collectively underscore the superior efficacy of MeshConv3D in comparison to alternative existing solutions incorporating both convolutional and pooling layers.

MeshConv3D exhibits notable advantages in terms of memory utilization and computational efficiency, while maintaining competitive performance in classification across various databases. Such outcomes not only highlight the promising potential of MeshConv3D in mesh data processing tasks but also emphasize its practical viability for real-world applications.

V. CONCLUSIONS AND FUTURE WORK

In this paper, we have introduced the MeshConv3D methodology, which includes a novel convolutional operator designed to adapt to varying kernel sizes, accompanied by an innovative pooling technique customized for learning on triangular meshes. MeshConv3D demonstrates competitive performance in semantic classification tasks, surpassing the current state-of-the-art benchmarks of models integrating both pooling and convolutional layers. Notably, the proposed methodology showcases a high efficiency in terms of time and GPU memory consumption, rendering it highly suitable for real-world applications.

Our future work will concern the development of further refinements and optimizations, related notably to the specification of more advances convolutional kernels that can seamlessly integrate within a unified framework salient geometric and topological features.